\documentclass[letterpaper]{article} 
\usepackage{aaai24}  
\usepackage{times}  
\usepackage{helvet}  
\usepackage{courier}  
\usepackage[hyphens]{url}  
\usepackage{graphicx} 
\usepackage{natbib}  
\usepackage{caption} 
\usepackage{enumitem}
\usepackage{algorithm}
\usepackage{algorithmic}
\usepackage{amsmath,amssymb,amsfonts}
\usepackage{import}
\usepackage{textcomp}
\usepackage{mathtools}
\usepackage{adjustbox}
\usepackage{multirow}
\usepackage{multicol}
\usepackage{makecell}
\usepackage{booktabs}
\usepackage{pifont}
\usepackage{dsfont}
\usepackage{subfigure}
\usepackage{xcolor}
\usepackage{arydshln}

\definecolor{brightpink}{rgb}{1.0, 0.0, 0.5}
\definecolor{cyan}{rgb}{0.0, 1.0, 1.0}
\definecolor{lightskyblue}{rgb}{0.53, 0.81, 0.98}
\definecolor{deepskyblue}{rgb}{0.0, 0.75, 1.0}

\usepackage{newfloat}
\usepackage{listings}
\DeclareCaptionStyle{ruled}{labelfont=normalfont,labelsep=colon,strut=off} 
\floatstyle{ruled}
\newfloat{listing}{tb}{lst}{}
\floatname{listing}{Listing}
\title{iDet3D: Towards Efficient Interactive Object Detection for LiDAR Point Clouds}
\author{
    Dongmin Choi\equalcontrib \quad
    Wonwoo Cho\equalcontrib \quad
    Kangyeol Kim \quad
    Jaegul Choo
}
\affiliations{
    Letsur Inc. \quad
    KAIST \\
    \{dmchoi, wcho, kangyeolk, jchoo\}@kaist.ac.kr
\\
\setcounter{figure}{0}
\vspace{9mm}
\begin{center}
\centering
\captionsetup{type=figure}
\includegraphics[width=0.95\textwidth]{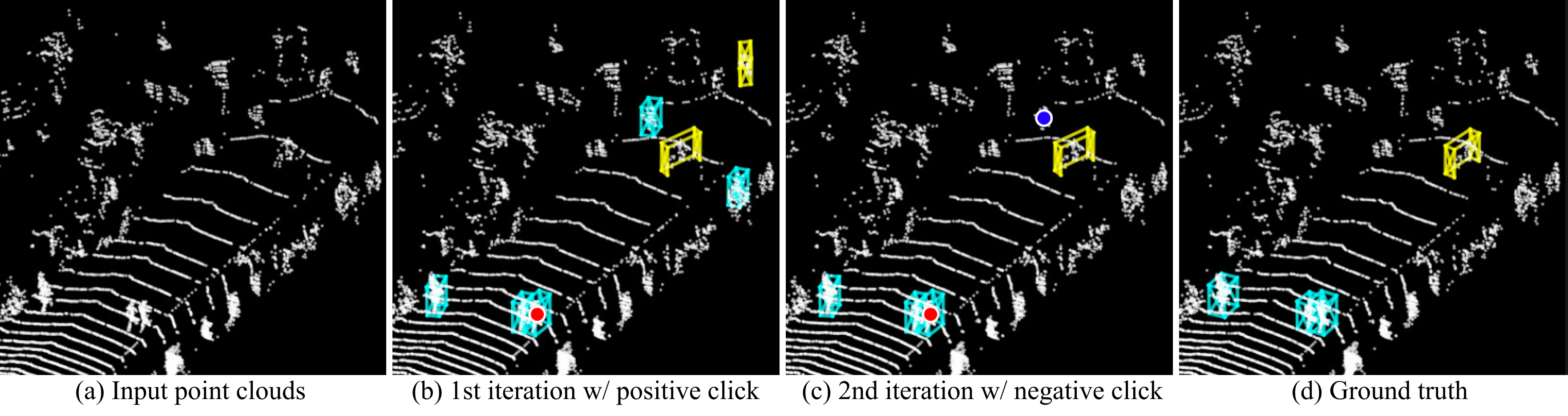}
     \captionof{figure}{
         An example of the iterative annotation process of iDet3D.
         (a) Given input point clouds.  
         (b) Provided a positive click on a pedestrian (\textcolor{red}{red} circle), the proposed iDet3D detects multiple objects of various classes in the scene within a single click.
         (c) In the second iteration, the false-positive predictions can be filtered out at once by adding a single negative click
         (\textcolor{blue}{blue} circle).
         (d) Ground truth.
          Within a few iterations, one can obtain high-quality annotation results.
 }
\label{fig:qualitative_iter}
\end{center}%
\vspace{4mm}

}

\begin{document}
\maketitle


\begin{abstract}
Accurately annotating multiple 3D objects in LiDAR scenes is laborious and challenging. While a few previous studies have attempted to leverage semi-automatic methods for cost-effective bounding box annotation, such methods have limitations in efficiently handling numerous multi-class objects. To effectively accelerate 3D annotation pipelines, we propose \textbf{iDet3D}, an efficient interactive 3D object detector. Supporting a user-friendly 2D interface, which can ease the cognitive burden of exploring 3D space to provide click interactions, iDet3D enables users to annotate the entire objects in each scene with minimal interactions. Taking the sparse nature of 3D point clouds into account, we design a negative click simulation (NCS) to improve accuracy by reducing false-positive predictions. In addition, iDet3D incorporates two click propagation techniques to take full advantage of user interactions: (1) dense click guidance (DCG) for keeping user-provided information throughout the network and (2) spatial click propagation (SCP) for detecting other instances of the same class based on the user-specified objects. Through our extensive experiments, we present that our method can construct precise annotations in a few clicks, which shows the practicality as an efficient annotation tool for 3D object detection.
\end{abstract}

\begin{figure*}[t] 
    \centering
        \includegraphics[width=1.0\textwidth]{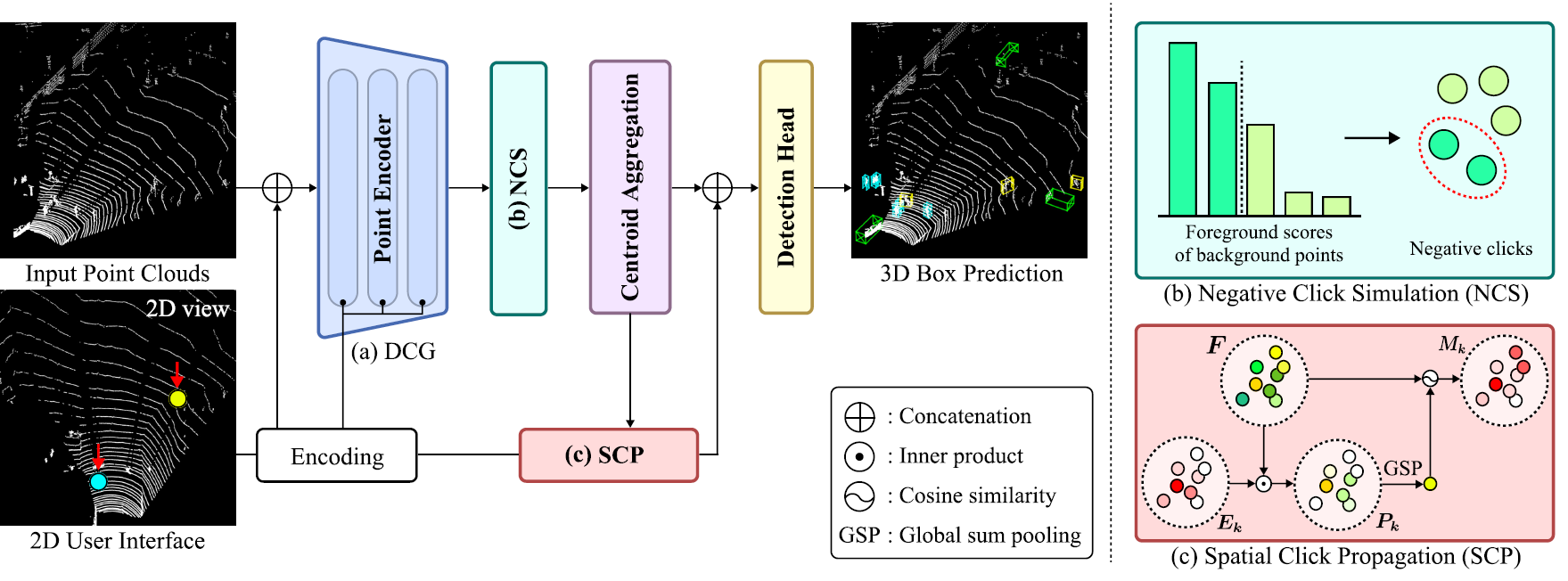}
        \caption{
            The training workflow of iDet3D.
            Given user clicks on target objects, the clicks are transformed into click encodings.
            (a) Dense click guidance (DCG) fuses the encodings into the backbone network architecture not only at the input side but also at the intermediate layers.
            (b) Negative click simulation (NCS) randomly simulates probable negative points by selecting challenging background points with high foreground scores.   
            (c) The following spatial click propagation (SCP) module effectively propagates user clicks to detect other objects of the same class based on the similarity between feature embeddings.
        }
        \label{fig:idet3d}
\end{figure*}

\section{Introduction}

3D object detection is a long-standing research topic that has been actively studied in industrial fields such as autonomous driving~\cite{geiger2012we,mao20223d} and robotics~\cite{geiger2013vision,wang2015voting}.
Although LiDAR point clouds have been widely used to efficiently represent complex scenes in 3D applications, their sparse and orderless nature leads their annotation process to be costly and erroneous~\cite{luo2023exploring,hu2022lidal,wu2021redal}, which can be a bottleneck for developing robust 3D object detectors.
For instance, even the renowned benchmark dataset, KITTI~\cite{geiger2012we}, contains several mislabeled objects~\cite{li2020sustech}.



To alleviate the complexity of LiDAR object annotation,
this paper incorporates \textit{user interactions} (\textit{i.e.,} clicks) into 3D object detectors as in the literature of interactive segmentation~\cite{jang2019interactive,lin2020interactive,liu2022ss3d}.
However, due to the unique aspects of point cloud data,
interactive annotation methods should consider the following requirements. 
(1) Distinguishing the points of foreground objects from the others is challenging in sparse point clouds~\cite{guo2020deep}; \emph{mispredictions should be effectively handled.}
(2) A point cloud scene often contains multiple 3D instances of different categories; \emph{each user interaction should not be limited to a single object}.
Thus, without properly taking account of the aspects, simply extending existing interactive approaches to 3D object detection may produce sub-optimal and insufficient results.


By addressing the aforementioned requirements, we propose an interactive 3D object detector called \textbf{iDet3D}, which includes
three components: negative click simulation (NCS), dense click guidance (DCG), and 
correlation-based spatial click propagation (SCP).
Based on our user-friendly annotation and click-encoding algorithm for point clouds,
which allow users to easily click on a 2D visual interface without laboriously exploring 3D space, 
such components boost the efficiency and effectiveness of our proposed iDet3D.





Since iDet3D should be able to effectively remove false-positive predictions by applying a few negative clicks, it is important to simulate negative clicks at training time properly.
Our NCS strategy aims to assign negative click candidates on background points that are prone to be mistakenly predicted as foreground instances.
In addition, the two click propagation strategies, DCG and SCP,
helps iDet3D to maintain and propagate user guidance throughout deep network layers and spatial point embeddings, respectively.
To the best of our knowledge, iDet3D, which is supported by the proposed components, is the first interactive 3D framework capable of detecting numerous 3D multi-class objects.

In Fig.~\ref{fig:qualitative_iter},
we demonstrate an interactive annotation example of our proposed iDet3D,
giving a positive click on the pedestrian followed by a negative click to suppress the false positives. 
Our main contributions are as follows:
\begin{itemize}[topsep = 0pt, itemsep=0pt]
    \item We propose iDet3D, a novel interactive 3D object detector that detects multiple objects of different categories in LiDAR point clouds within a few user clicks.

    \item The NCS strategy makes iDet3D be capable of reducing false-positives by leveraging user-given negative clicks.
    
    \item We carefully design effective click propagation methods (DCG and SCP) to take full advantage of user-provided interactions throughout the network and a 3D scene.
    
    \item Our extensive experiments on several 3D LiDAR datasets shows the effectiveness of iDet3D as an annotation tool.
\end{itemize}

\section{Related Work}
\subsection{3D Object Detectors}

\paragraph{Voxel-based.}

To use the mature convolutional neural networks (CNNs) in processing the unordered 3D point clouds, previous studies~\cite{he2020structure,lang2019pointpillars,liu2020tanet,shi2020point,zheng2021cia} have proposed to transform sparse point clouds into dense voxel grids through spatial quantization (\emph{i.e.,} voxelization).
Although this enables CNN-based analysis of 3D scenes~\cite{wang2015voting} via 3D sparse convolution~\cite{yan2018second} or pillars~\cite{lang2019pointpillars}, voxel-based methods inevitably suffer from information loss caused by their inability to fully exploit the original structural information~\cite{xu2021rpvnet,akhtar2020point}.
In addition, their computational costs increase cubically with input resolution, 
while requiring additional time for input pre-processing (voxelization) and post-processing (devoxelization).

In interactive applications, it is intuitive that the response time (or latency) of an algorithm is particularly important for a better user experience.
Since the aforementioned problems of voxel-based detectors may hinder flexible real-time interactions,
we consider point-based backbones in this paper.

\paragraph{Point-based.}
Unlike voxel-based ones, point-based detectors directly take 3D point coordinates as input and then analyze per-point embeddings.
After PointNet~\cite{qi2017pointnet}, the first backbone architecture that directly analyzes point clouds, was introduced,
various point-based detectors~\cite{shi2020point,li20203d,yang20203dssd,zhang2022not} have been developed. Compared to voxel-based models, point-based methods are generally more efficient in terms of memory~\cite{guo2020deep}.

Point-based detectors can be divided into \emph{two-stage} and \emph{single-stage} detectors. Although early studies have focused on two-stage approaches~\cite{shi2019pointrcnn, qi2019deep} due to the limited detection accuracy of single-stage detectors, 
two-stage detectors can also suffer from low inference speed problems caused by their time-consuming layers.
Thus, recent studies have focused on designing efficient but effective single-stage 3D detectors.
3DSSD~\cite{yang20203dssd} was proposed as the first single-stage point-based 3D object detector, which removes computational-heavy components for 3D proposals and overcomes the following performance drop through semantic feature-based farthest point sampling (FPS).
Afterwards, a highly efficient 3D detector called IA-SSD~\cite{zhang2022not} was introduced for deployment in practice, which replaces time-consuming FPS with instance-aware downsampling MLP layers.

Since IA-SSD achieved competitive detection accuracy while maintaining high efficiency, we employ this architecture as the backbone of iDet3D.
To present that the principle of iDet3D is applicable to other single-stage detectors, we also exploit 3DSSD backbone in our experiment.

\subsection{Interactive Point Cloud Annotation}

A few approaches have been proposed to incorporate interactive techniques into 3D point clouds.
For point cloud segmentation tasks,
scribble and click-based interactive refinement approaches have been investigated~\cite{shen2020interactive,kontogianni2022interactive}.
In 3D object detection,
an interactive annotation framework based on LiDAR sensor fusion and one-click bounding box drawing was introduced~\cite{wang2019latte}.
Afterwards, a more advanced annotation system that supports smart 3D bounding box initialization and automatic box fitting was developed~\cite{li2020sustech}.

While the previous methods can accelerate point cloud annotation compared to manual labeling processes,
such interactive schemes are still limited to identifying an individual instance at once, \emph{i.e.,}
users can modify the annotation of only a single object for each interaction.

\section{Method}


\paragraph{Overview.}

Throughout this paper, we describe our proposed iDet3D based on the IA-SSD backbone~\cite{zhang2022not}, a recently proposed 3D object detector.
It is noteworthy that our principle can be easily applicable to
other single-stage point-based detectors.
iDet3D supports two types of user interactions: \emph{class-specific} positive and \emph{class-agnostic} negative clicks, which are designed to indicate the locations of foreground objects and background regions, respectively.
Fig.~\ref{fig:idet3d} shows the overall architecture of iDet3D.

\subsection{Click Encoding}
\label{sec:vanilla}

A straightforward approach to provide interaction to a given 3D scene is to directly click
on the objects of interest~\cite{kontogianni2022interactive}.
However, the process of specifying the 3D coordinates of a small point in a vast 3D space imposes a significant cognitive burden on users.

Instead, we develop a user-friendly interface in 2D view, where users can provide simple 2D clicks on target objects.
For a better understanding, we visualize the difference between 3D and our 2D interfaces.
In the 3D interface, a slight shift of a cursor may lead to undesired movements of coordinates in another axis.
However, our 2D annotation environment can mitigate such errors by eliminating the requirement of specifying z-axis locations.

\begin{figure}[t] 
    \centering
    \includegraphics[width=0.48
    \textwidth]{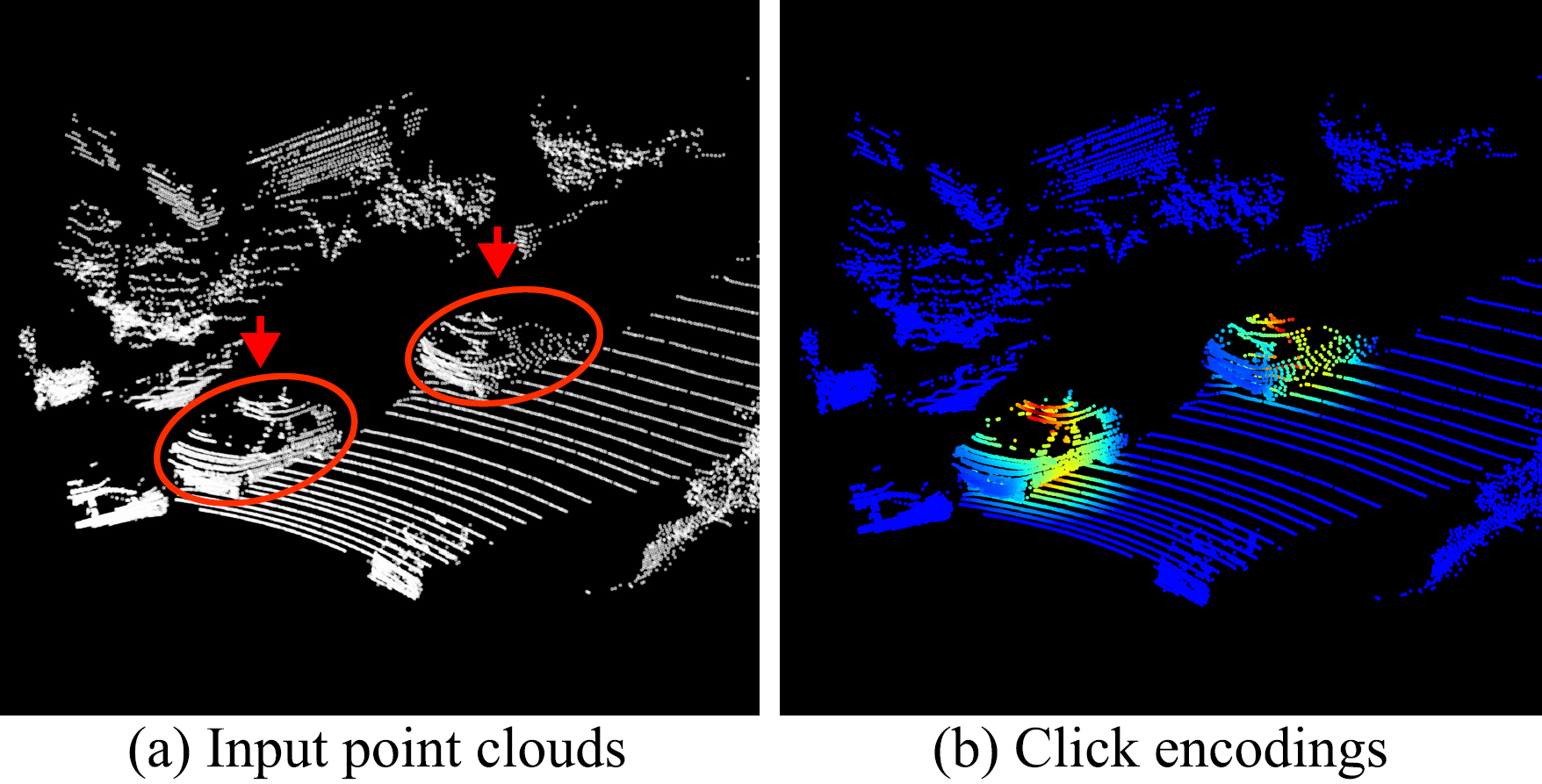}
    \caption{
        An example visual illustration of click encoding in our iDet3D.
        (a) Input point clouds and user clicks (\textcolor{red}{red} arrows).
        (b) The corresponding distance-encoded user interactions
        highlighted on the target objects.
    }
    \label{fig:encoding}
    \vspace{-1\baselineskip}
\end{figure}

\begin{figure*}[t] 
\centering
\includegraphics[width=0.99\textwidth]{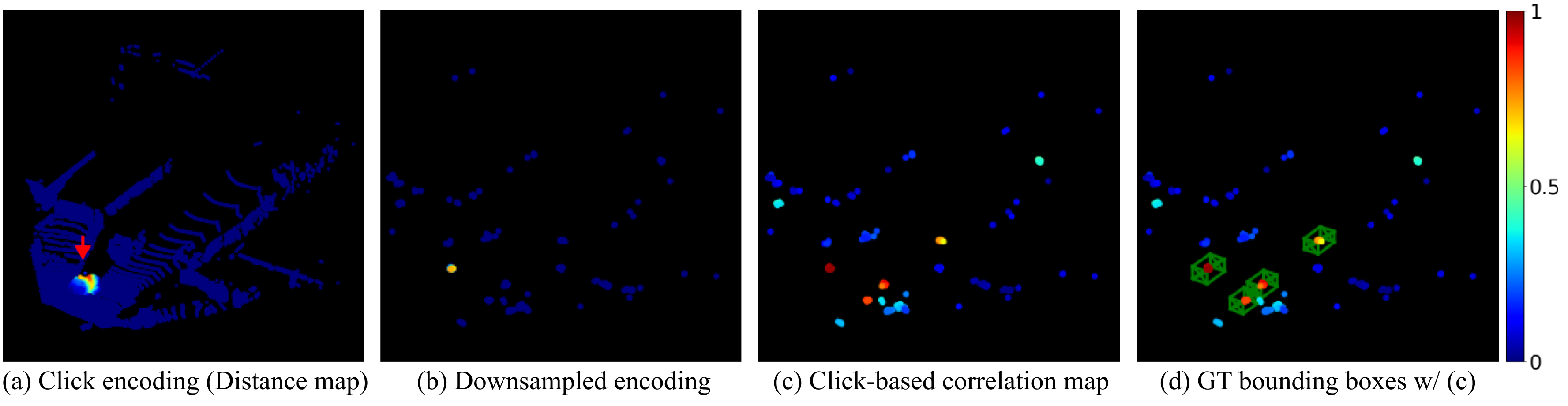}
\caption{
    An example of correlation map generation using spatial click propagation (SCP) module (visualized in the scale of [0,1]).
    (a) Click encoding $E$ on a car object (\textcolor{red}{red} arrow) with respect to $N$ input points.
    (b) Click encoding $E'$ for $N'$ downsampled points (after point reduction from $N$ to $N'$ with downsampling layers).
    (c) Click-wise correlation map $M$ generated by the SCP and
    (d) visualization overlaid with the ground-truth 3D bounding boxes.
    Note that points of the same class with the click are highlighted in the output correlation map.
}
\label{fig:scp}
\end{figure*}

Suppose that $K$ class-specific positive clicks are provided on a scene to annotate foreground objects of total $C$ categories.
Then, the $k$-th click can be written as $(p_k, c_k)$, where $p_k=(p_{k,x}, p_{k,y})$ denotes the 2D coordinate of the click, and $c_k \in \{1,\cdots,C\}$ is the corresponding class.
Following the convention of deep interactive annotation methods~\cite{xu2016deep},
we transform user clicks into the corresponding distance heatmap
to generate a proper input for point-based detectors.
Given a 3D point cloud scene composed of $N$ points $\{(x_i, y_i, z_i)\}_{i=1}^N$, we encode $(p_k, c_k)$ into a \emph{click-wise} encoding $E_k \in \mathbb{R}^{N}$, whose $i$-th element is
\begin{equation}
    E_k[i]=
    \exp \left({\max \left\{\frac{\tau-d} {\tau}, 0 \right\} \cdot \log{2}} \right) - 1.
\label{eq:clk_encoding}
\end{equation}
In Eq.~\eqref{eq:clk_encoding}, $d=\sqrt{(p_{k,x}-x_i)^2 + (p_{k,y}-y_i)^2}$ represents the 2D Euclidian distance between $p_k$ and $(x_i, y_i)$, and $\tau$ is a hyperparameter to control the distance threshold.
Note that $E_k$ is designed to highlight $p_k$ and its neighboring points within the $[0,1]$ scale.

To effectively feed the encoded clicks $\{E_k\}_{k=1}^K$ to networks,
we define a \emph{class-wise} click encoding $U_c \in \mathbb{R}^{N}$ for class $c \in \{1,\cdots,C\}$ via element-wise max pooling, \emph{i.e.,}
\begin{equation}
    U_c[i] = \max\{E_k[i] | c_k = c, \forall k \in \{1,\cdots,K\} \}.
\label{eq:cls_encoding}
\end{equation}
Once $C$ encodings are generated, we concatenate $\{U_c\}_{c=1}^C$ to the corresponding input points.
For better understanding, we visualize an example of $U_c$ computed by two clicks of Car class in Fig.~\ref{fig:encoding}.
We define a \emph{vanilla model} by a combination of this click encoding and the backbone encoder.

\subsection{Negative Click Simulation}
\label{sec:negative}

We observe that the vanilla model with only positive clicks fails to separate background point clouds from foreground ones, causing unexpected false-positive predictions.
To mitigate a similar problem, previous studies on interactive segmentation~\cite{xu2016deep,sofiiuk2022reviving} have made use of \emph{negative clicks} to indicate the undesired region.
In general, they randomly sample negative clicks based on the assumption that real users are likely to provide negative clicks to areas outside foreground regions but near object boundaries.
However, because \emph{false positives in 3D object detection can happen regardless of foreground object locations}, the simulation strategy of interactive segmentation may not derive reasonable negative clicks.

Instead, we propose negative click simulation (NCS) suitable for 3D object detection with the goal of sampling challenging background points that are likely to be inaccurately predicted as foreground.
For this purpose, we take advantage of MLP-based scoring embedded in the down-sampling approaches of recent point-based detectors~\cite{zhang2022not,chen2022sasa}.
This method assigns high scores to potential foreground points and selects top-$n$ points to be downsampled, which implies that several challenging background points can be ranked in top-$n$.

We expand the functionality of this layer to act as a negative click simulator by selecting \emph{background points with high foreground scores} as negative clicks.
After this simulation strategy, we sample top-$K_n$ background points and encode them with the same manner of positive clicks,
\emph{i.e.}, click encoding becomes $(C+1)$ channels, where the additional single channel is for class-agnostic negative clicks.

\begin{table*}[ht]
\begin{center}
\fontsize{6.}{9pt}\selectfont
\renewcommand{\arraystretch}{1.0}
\setlength{\doublerulesep}{0.9pt}
\begin{adjustbox}{width=1.0 \textwidth}
\begin{tabular}{c|c|c|ccc|ccc|ccc}
\hline
\multirow{2}{*}{} & \multirow{2}{*}{Method} & \multirow{2}{*}{$N_{\text{clks}}$} & \multicolumn{3}{c|}{3D Car (IoU=0.7)} & \multicolumn{3}{c|}{3D Ped. (IoU=0.5)} & \multicolumn{3}{c}{3D Cyc. (IoU=0.5)} \\
 & & & \textit{Easy} & \textit{Mod.} & \textit{Hard} & \textit{Easy} & \textit{Mod.} & \textit{Hard} & \textit{Easy} & \textit{Mod.} & \textit{Hard} \\
\hline

\multirow{4}{*}{\rotatebox[origin=c]{90}{Voxel}} & VoxelNet~\cite{zhou2018voxelnet} & - & 81.97 & 65.46 & 62.85 & 57.86 & 53.42 & 48.87 & 67.17 & 47.65 & 45.11 \\ 
 & SECOND (IoU)~\cite{yan2018second} & - & 84.88 & 76.30 & 75.97 & 37.26 & 34.60 & 32.62 & 80.63 & 64.35 & 60.38 \\
 & PointPillars~\cite{lang2019pointpillars} & - & 86.45 & 77.29 & 74.65 & 57.76 & 52.30 & 47.91 & 80.05 & 62.73 & 59.67 \\
 & Part-$A^{2}$ (Anchor)~\cite{shi2020points} & - & \underline{89.56} & 79.41 & \underline{78.84} & \underline{65.69} & \underline{60.05} & \underline{55.44} & 85.50 & 69.93 & 65.49  \\
\hline

\multirow{3}{*}{\rotatebox[origin=c]{90}{Point}}
 & 3DSSD (Reproduced)~\cite{yang20203dssd} & - & 89.12 & \underline{83.94} & 78.47 & 60.65 & 56.05 & 52.19 & 84.75 & 69.14 & 64.58 \\
 & IA-SSD~\cite{zhang2022not} & - & 89.47 & 79.57 & 78.45 & 62.38 & 58.91 & 51.46 & 86.65 & \underline{71.24} & 66.11 \\
 & IA-SSD (Reproduced)~\cite{zhang2022not} & - & 89.20 & 79.28 & 78.15 & 60.92 & 57.77 & 51.66 & \underline{87.16} & 68.25 & \underline{66.77} \\
\hline\hline

\multirow{12}{*}{\rotatebox[origin=c]{90}{Interactive}} 
    & \multirow{3}{*}{\makecell{Vanilla iDet3D \\ (3DSSD backbone)}} & 1 & 88.82	& 83.87 & 78.38 & 62.56 & 57.36 & 53.45 & 84.35 & 67.42 & 64.50 \\
 & & 3 & 88.85 & 83.93 & 78.42 & 62.72 & 58.29 & 53.81 & 84.85 & 68.65 & 64.86 \\
 & & 5 & 88.86 & 83.94 & 78.42 & 62.89 & 58.60 & 53.90 & 84.83 & 68.18 & 64.84 \\
\cdashline{2-12}[1.5pt / 2pt]
    & \multirow{3}{*}{\makecell{Vanilla iDet3D \\ (IA-SSD backbone)}} & 1 & 88.70 & 79.28 & 78.34 & 58.71 & 55.47 & 50.98 & 86.47 & 70.40 & 65.49 \\
 & & 3 & 88.76 & 79.30 & 78.36 & 58.78 & 55.67 & 51.07 & 86.49 & 70.45 & 65.58 \\
 & & 5 & 88.76 & 79.31 & 78.37 & 58.76 & 55.72 & 51.07 & 86.51 & 70.44 & 65.56 \\
\cline{2-12}
  & \multirow{3}{*}{\makecell{iDet3D \\ (3DSSD backbone)}} & 1 & 91.17 & 88.09 & 86.79 & 60.06 & 57.66 & 52.21 & 88.14 & 73.21 & 69.29 \\
 & & 3 & 96.26 & 88.91 & 88.51 & 62.26 & 58.79 & 55.08 & 88.88 & 75.73 & 72.66 \\
 & & 5 & 97.15 & 88.96 & 88.61 & 62.93 & 59.08 & 55.28 & 88.92 & 76.06 & 73.09 \\
\cdashline{2-12}[1.5pt / 2pt]
  & \multirow{3}{*}{\makecell{iDet3D \\ (IA-SSD backbone)}} & 1 & 89.83 & 87.99 & 85.68 & 60.87 & 57.27 & 51.66 & 90.60 & 74.69 & 73.00 \\
 & & 3 & 97.21 & 89.74 & 89.47 & 66.13 & 62.37 & 57.40 & 96.46 & 83.31 & 77.91 \\
 & & 5 & \textbf{98.55} & \textbf{90.37} & \textbf{90.27} & \textbf{70.07} & \textbf{65.74} & \textbf{60.57} & \textbf{98.43} & \textbf{88.00} & \textbf{80.19} \\

\hline
\end{tabular}
\end{adjustbox}
\end{center}
\caption{
    Quantitative results of baselines and iDet3D on the KITTI \textit{val} set.
    The best results (measured in the AP metric) among the non-interactive models are underlined, where the best ones of the interactive methods are highlighted in bold.
}
\label{table:kitti_val}
\end{table*}

\subsection{User Click Propagation}
\label{sec:propagation}
In addition to the limitation of false-positive predictions, we discover that the vanilla model sometimes fails to detect the user-specified object.
This finding implies two drawbacks of the model: 
(1) \emph{user intention can be diluted through the forward pass of backbone layers} and (2) \emph{user clicks are limited in affecting multiple objects}.
To address these problems, we propose two click propagation methods, which are DCG to make iDet3D sustain user intention and SCP to empower a user click to influence other objects of the same category.

\paragraph{Dense click guidance (DCG).}
If the user click encoding $\{U_c\}_{c=1}^{C+1}$
are only fused to the input points, user intention in the click guidance can be diluted as the network layer goes deeper~\cite{zhang2019late,ding2022rethinking,hao2021edgeflow}.
Furthermore, the point-based detectors with downsampling layers for computational efficiency may cause potential losses of foreground points or critical information for 3D scene understanding~\cite{hu2021learning,zhang2022not}.
Thus, the following prediction head may be unable to effectively leverage user-provided hints.

To address this problem, we concatenate encoded clicks to input point clouds as well as the intermediate point embeddings of the encoder after each downsampling layer, as illustrated in Fig.~\ref{fig:idet3d}(b).
This \emph{dense guidance} strategy greatly helps iDet3D to utilize user guidance throughout the network without forgetting the user intention.

\paragraph{Spatial click propagation (SCP).}

Most 3D scenes contain not a single but multiple objects.
For better efficiency, it is required to utilize user guidance to detect all instances including unspecified objects.
However, the click encoding obtained by following Eq.~\eqref{eq:clk_encoding} only highlights the single object specified by users.
Therefore, it is limited to affecting other instances that have not been explicitly indicated.

Inspired by object counting algorithms~\cite{arteta2014interactive,ranjan2021learning} and a multi-class 2D interactive detector~\cite{lee2022interactive},
we add the SCP module to the output of the encoder
to enhance the click efficiency (Fig.~\ref{fig:idet3d}).
Let the output point embeddings of the encoder be $F\in \mathbb{R}^{N' \times D}$, where $N'$ is the number of downsampled points and $D$ indicates the dimension of the embedding.
To be aligned with the downsampled points, we define $E_k'\in \mathbb{R}^{N'}$, a downsampled $E_k$ for the $k$-th click, by recomputing Eq.~\eqref{eq:clk_encoding} with respect to the encoder output features $F$.
Then, we compute a click prototype vector $P_k \in \mathbb{R}^{D}$ by
\begin{equation}
    P_k = \sum_{j}^{N'} \left( F[j,:] \cdot \frac{E_k'[j]}{\lVert E_k'\lVert_1} \right),
\label{eq:template}
\end{equation}
where $F[j,:] \in \mathbb{R}^{D}$ is the $j$-th point embedding and $\lVert E_k'\lVert_1$ is the $L_1$-norm of $E_k'$.
In other words, each $P_k$ is a weighted sum of point embeddings corresponding to the neighboring points of the click, which encodes the prototype representation of the object indicated by the $k$-th click.


For $P_k$, we compute a \emph{click-wise} correlation map
$M_k \in \mathbb{R}^{N'}$ based on the cosine similarity followed by a global sum pooling, which can be represented as
\begin{equation}
    M_k[j] = \frac{F[j, :] \odot P_k}
        {\lVert F[j, :] \rVert_2 \lVert P_k \rVert_2},
\label{eq:classwise_sim}
\end{equation}
where $F[j, :] \odot P_k$ is the inner product of $F[j, :]$ and $P_k$.
As $M_k$ is designed to highlight those points with high cosine similarity between their feature embeddings and $P_k$, 
it is able to spatially propagate a click on a single object to other unspecified ones belonging to the same class.
To provide further insight, we illustrate an example in Fig.~\ref{fig:scp}.

To aggregate the given click-wise maps before incorporating them into the network, 
we compute a class-wise correlation map $S_c \in \mathbb{R}^{N'}$ for the class $c$ by
\begin{equation}
    S_c[j] = \max\{M_k[j] | c_k = c, \forall k \in \{1,\cdots,K\} \}.
\end{equation}

Following both DCG and SCP, point embeddings (with 
class-wise click encodings of $C+1$ dimension and correlation maps of $C+1$ dimension) become $(D+2C+2)$-dimension vectors, and the detection head takes them as input to make final predictions.
It is noteworthy that negative clicks in our system can also affect other unspecified background points since the SCP module generates a negative click-based correlation map.

\begin{figure*}[t] 
\centering
\includegraphics[width=1.0\textwidth]{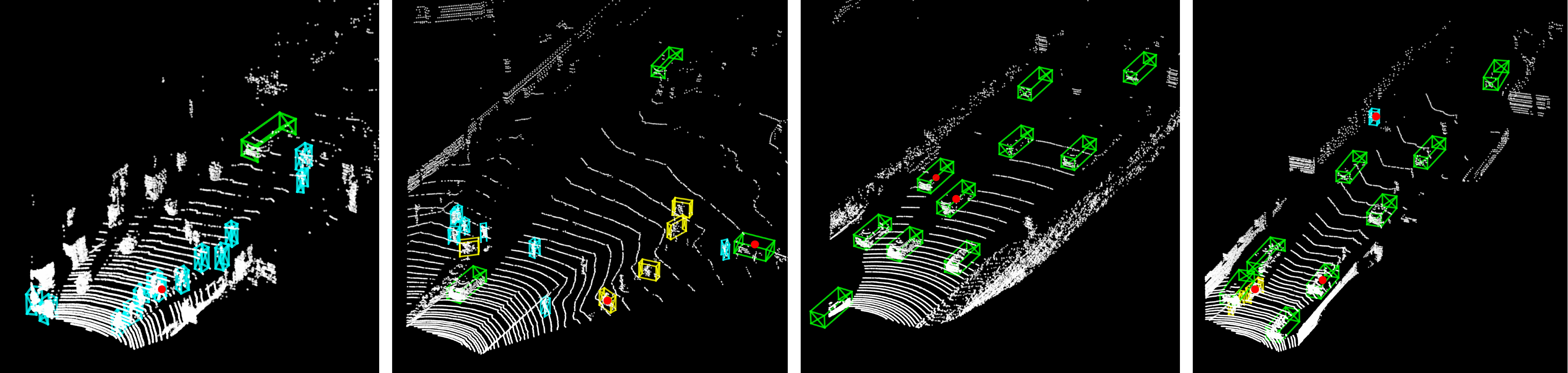}
\caption{
    Qualitative results of iDet3D on the KITTI \emph{val} set.
    \textcolor{green}{Green} boxes for Car, \textcolor{cyan}{cyan} for Pedestrian, and \textcolor{yellow}{yellow} for Cyclist with user clicks in \textcolor{red}{red} circles.
    Our model successfully detects multiple objects of different categories with a few user clicks.
}
\label{fig:kitti_qualitative}
\end{figure*}

\begin{figure*}[t] 
\centering
\includegraphics[width=1.0\textwidth]{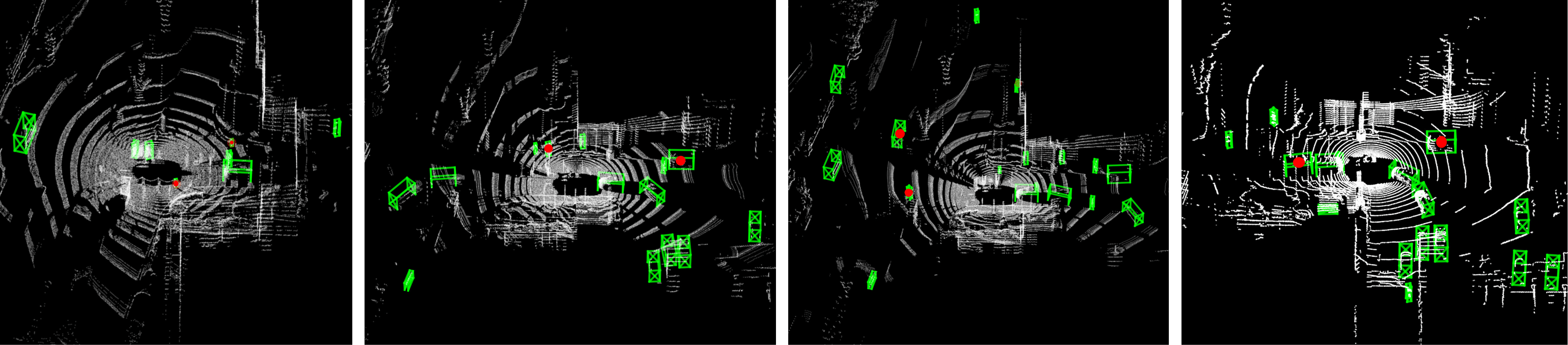}
\caption{
    Qualitative results of iDet3D on the nuScenes validation set.
    The 3D bounding box predictions are colored with \textcolor{green}{green},
    and the user clicks are represented as \textcolor{red}{red} circles.
}
\label{fig:nuscenes_qualitative}
\end{figure*}

\section{Experiments}

\subsection{Experimental Settings}

\paragraph{Evaluation protocols.}

We evaluate iDet3D with comparison to automatic (non-interactive) baselines including voxel-based~\cite{zhou2018voxelnet,yan2018second,wang2020pillar,liu2020tanet,shi2020points} and point-based 3D object detectors~\cite{yang20203dssd,zhang2022not}.

Furthermore, the \textit{vanilla} interactive model (without NCS, DCG, and SCP) serves as a simple baseline.
Following the convention of evaluation protocol in interactive segmentation~\cite{li2018interactive,sofiiuk2020f,sofiiuk2022reviving}, we measure the performance by iteratively increasing the number of clicks.

At each interaction, we prioritize reducing false-negative predictions by adding a positive click.
If all target objects in a given 3D scene are predicted, we then provide a negative click to suppress false-positive cases if they exist.
Quantitative results are reported by averaging the scores of five randomized click sampling trials.

\paragraph{Datasets.}

KITTI benchmark~\cite{geiger2012we} is a widely used 3D object detection dataset, which consists of 3,712 training and 3,769 validation samples with three object classes: Car, Pedestrian, and Cyclist.
Following the official KITTI evaluation protocol, we measure the average precision (AP) metric with an intersection over union (IoU) threshold of 0.7 for Car and 0.5 for Pedestrian and Cyclist.
For evaluation, we use KITTI \emph{val} set instead of the \emph{test} set because a user click simulation is not allowed in benchmark submission due to the unavailability of ground truth annotation.
We also evaluate iDet3D on a more challenging nuScenes dataset~\cite{caesar2020nuscenes}, which contains 1,000 scenes recorded in Boston and Singapore comprising 20,000 frames.
In total, nuScenes includes approximately 1.4M objects with 10 object categories.

\paragraph{Implementation details.}
During training and evaluation, we perform positive click simulation by randomly sampling 2D coordinates inside the 3D ground-truth boxes.
The number of clicks $K$ is determined as $\min{(N_u, N_o)}$, where $N_u$ is sampled from $\{0, \cdots, 10\}$ uniformly at random and $N_o$ refers to the number of existing objects in each scene.
The distance threshold $\tau$ of user encodings is set to 2.0 in Eq.~\eqref{eq:clk_encoding}.
For negative clicks, we set the maximum number $K_n$ to 10.

For more detailed training configurations for each backbone architecture, readers are referred to
the codebases of IA-SSD~\cite{zhang2022not} and SASA~\cite{chen2022sasa}.
We use 4 NVIDIA RTX A6000 GPUs for experiments.



\begin{table*}[ht!]
\begin{center}
\setlength{\doublerulesep}{0.9pt}
\fontsize{6.}{7.5pt}\selectfont
\renewcommand{\arraystretch}{1.1}
\begin{adjustbox}{width=1.0\textwidth}
\begin{tabular}{c|c|c|c|c|c|c|c|c|c|c|c|c|c}
\hline
 & Method & $N_{\text{clks}}$ & Car & Ped & Bus & Barrier & T.C. & Truck & Trailer & Moto & C.V. & Bicycle & mAP \\
\hline

\multirow{2}{*}{Voxel-based} & PointPillars~\cite{lang2019pointpillars} & - & 70.5 & 59.9 & 34.4 & 33.2 & 29.6 & 25.0 & 20.0 & 16.7 & 4.5 & 1.6 & 29.5  \\
 & SECOND~\cite{yan2018second} & - & 75.53 & 59.86 & 29.04 & 32.21 & 22.49 & 21.88 & 12.96 & 16.89 & 0.36 & 0 & 27.12  \\
\hline

\multirow{3}{*}{Point-based} & 3DSSD~\cite{yang20203dssd} & - & \underline{81.20} & \underline{70.17} & 61.41 & 47.94 & \underline{31.06} & \underline{47.15} & 30.45 & 35.96 & 12.64 & 8.63 & 42.66 \\
 & SASA~\cite{chen2022sasa} & - & 76.8 & 69.1 & 66.2 & \underline{53.6} & 29.9 & 45.0 & \underline{36.5} & \underline{39.6} & 16.1 & \underline{16.9} & \underline{45.0} \\
 & IA-SSD*~\cite{zhang2022not} & - & 72.84 & 61.51 & \underline{66.22} & 37.02 & 22.08 & 41.63 & 29.26 & 34.95 & \underline{17.93} & 11.76 & 39.51 \\
 
\hline\hline

\multirow{6
}{*}{Interactive} 
& \multirow{3}{*}{Vanilla iDet3D} & 1 & 72.63 & 60.93 & 65.78 & 37.35 & 21.93 & 41.51 & 26.97 & 34.01 & 17.96 & 12.68 & 39.17  \\
 & & 3 & 72.64 & 61.16 & 65.77 & 37.61 & 22.17 & 41.57 & 26.96 & 34.68 & 17.93 & 12.93 & 39.34 \\
 & & 5 & 72.64 & 61.39 & 65.72 & 37.80 & 22.34 & 41.57 & 26.96 & 35.22 & 17.93 & 13.17 & 39.48 \\
 \cline{2-14}
  & \multirow{3}{*}{iDet3D} & 1 & 73.41 & 62.99 & 66.97 & 39.59 & 22.96 & 43.31 & 26.74 & 39.97 & 21.41 & 19.03 & 41.64 \\
 & & 3 & 75.57 & 67.61 & 68.44 & 46.02 & 29.64 & 46.28 & 27.58 & 49.18 & 27.77 & 33.62 & 47.17 \\
     & & 5 & \textbf{77.28} & \textbf{71.14} & \textbf{69.70} & \textbf{50.68} & \textbf{35.71} & \textbf{48.67} & \textbf{29.28} & \textbf{56.19} & \textbf{34.51} & \textbf{45.16} & \textbf{51.83} \\
 
\hline
\end{tabular}
\end{adjustbox}
\end{center}
\caption{
    Quantitative results of the baselines and iDet3D (IA-SSD backbone) with the nuScenes dataset.
    The best results among the non-interactive models are underlined, where the best ones of the interactive methods are highlighted in bold.
    IA-SSD* is the reproduced version of~\cite{zhang2022not} by adapting the training configuration in the codebase of~\cite{chen2022sasa}.
}
\label{table:nuscenes_val}
\vspace{-1\baselineskip}

\end{table*}

\begin{table}[t]
\begin{center}
\renewcommand{\arraystretch}{1.2}
\begin{adjustbox}{width=0.49\textwidth}
\begin{tabular}{cccc|cc|cc|cc}
\hline
\multirow{2}{*}{} & \multirow{2}{*}{DCG} & \multirow{2}{*}{NCS} & \multirow{2}{*}{SCP} & \multicolumn{2}{c|}{Car (IoU=0.7)} & \multicolumn{2}{c|}{Ped. (IoU=0.5)} & \multicolumn{2}{c}{Cyc. (IoU=0.5)} \\
 & & & & \textit{Easy.} & \textit{Mod.} & \textit{Easy.} & \textit{Mod.} & \textit{Easy.} & \textit{Mod.} \\
\hline

(1) & - & - & - & 88.76 & 82.08 & 58.76 & 56.60 & 86.51 & 74.21 \\
(2) & \checkmark & - & - & 90.84 & 88.86 & 60.39 & 58.89 & 89.00 & 76.24 \\
(3) & \checkmark & \checkmark & - & 92.11 & 90.30 & 69.94 & 65.15 & \textbf{98.46} & 84.60 \\
(4) & \checkmark & \checkmark & \checkmark & \textbf{98.55} & \textbf{90.37} & \textbf{70.07} & \textbf{65.74} & 98.43 & \textbf{88.00} \\

 
\hline
\end{tabular}
\end{adjustbox}
\end{center}
\caption{
    Ablation study of iDet3D with five clicks.
    Adding DCG, NCS, and SCP consistently boosts detection accuracy.
}
\label{table:ablation}
\vspace{-1\baselineskip}
\end{table}




\subsection{Experimental Results}

In the experiments, our main interest is two-fold: (1) demonstrating the effectiveness of the interactive approach compared to state-of-the-art automatic (non-interactive) detectors,
\emph{i.e.,} the performance can be significantly improved by using a few user clicks, and (2) validating the effectiveness of the proposed components of iDet3D (NCS, DCG, and SCP) by comparing iDet3D to the vanilla model.

\paragraph{KITTI.} 

Table~\ref{table:kitti_val} shows quantitative comparison results between the baselines on the KITTI validation set.
As shown in the table, iDet3D achieves superior or competitive results compared to non-interactive baselines,
which implies that a few user clicks can be fulfilling sources for our model. 
It is noteworthy that even a single click can be effective in handling relatively challenging cases (\textit{Moderate} and \textit{Hard} cases of Car and Cyclist).
Also, as the number of clicks increases, the detection accuracy gradually improves, indicating that additional user clicks are successfully incorporated.
Several qualitative results are shown in Fig.~\ref{fig:kitti_qualitative}.

\paragraph{nuScenes.} 
Next, we perform an evaluation on nuScenes, which has a larger number of object classes than KITTI.
In comparison with our backbone IA-SSD~\cite{zhang2022not}, Table~\ref{table:nuscenes_val} shows that iDet3D achieves superior detection performance by applying only a single click per frame.
Although direct comparison between the detection accuracy of
our implemented IA-SSD (and iDet3D) on nuSences and the reported best accuracy of other non-interactive baselines can be misleading,
we confirm that iDet3D with five user clicks shows superior performance compared to the baselines in most of the categories and significantly outperforms them with respect to mAP.
Especially, user clicks effectively work on challenging classes
indicating that a few clicks can significantly aid our proposed model for 3D detection in LiDAR point clouds.
We visualize qualitative results in Fig.~\ref{fig:nuscenes_qualitative}.


\subsection{Additional Results}

\paragraph{Ablation study.}
Our ablation study analyzes the effect of each component of iDet3D based on the KITTI \emph{val}.
As reported in Table~\ref{table:ablation},
DCG leads to an overall improvement in detection accuracy.
This finding suggests that retaining click information throughout the feature extraction process in the encoder is critical.
Also, we also discover that NCS significantly enhances overall performance, which emphasizes the necessity of negative clicks and the effectiveness of our simulation method.
Lastly, a combination of all components enhances overall performance. 
It implies that the SCP module successfully propagates positive and negative clicks to the entire 3D scene,
thus enhancing our click efficiency.


\paragraph{Another backbone.}
To show that the principle of iDet3D can be applicable to another backbone architecture,
we also employ 3DSSD~\cite{yang20203dssd} as our backbone network.
For NCS, we adapt centerness values computed in 3DSSD as foreground scores. 
The corresponding experimental results based on the KITTI \emph{val.} set are presented in Table~\ref{table:kitti_val},
where the results present that our proposed components can be incorporated into other single-stage point-based backbones.

\paragraph{Comparison with another annotation framework.}

Furthermore, we quantitatively compare our iDet3D to
another semi-automatic annotation framework LATTE~\cite{wang2019latte}, where the system generates a single bounding box by receiving a click provided on each object (one-to-one).
As shown in Table~\ref{table:pr_latte}, our framework that detects multiple objects of different classes in a few clicks (many-to-many) outperforms LATTE in accuracy and click efficiency.

Furthermore, to supplement the simulation-based experimental results,
we conduct a user study with well-educated annotators. The study compares the efficiency between manual annotation, LATTE, and our proposed iDet3D in terms of the average number of clicks and time required to complete annotating each 3D scene
as shown in Table~\ref{table:user_study}.


\begin{table}[t]
\begin{center}
\renewcommand{\arraystretch}{1.2}
\begin{adjustbox}{width=0.38\textwidth}
\begin{tabular}{c|c|c|c}
\hline
Method & Precision (\%) & Recall (\%) & $N_\text{clks}$ / instance \\
\hline
LATTE & 78.8 & 84.8 & 1.29 \\
\hline

\multirow{2}{*}{iDet3D} & 82.7 & 85.9 & 0.23 \\
 & \textbf{83.8} & \textbf{88.0} & \textbf{0.69} \\
 
\hline
\end{tabular}
\end{adjustbox}
\end{center}
\caption{
    Comparison between iDet3D and LATTE.
    The last column shows the average number of clicks per instance.
}
\label{table:pr_latte}
\end{table}

\begin{table}[t]
\begin{center}
\renewcommand{\arraystretch}{1.2}
\begin{adjustbox}{width=0.42\textwidth}
\begin{tabular}{c|c|c}
\hline
Method & Avg. \# of clicks per scene & Avg. time per scene \\
\hline
 Manual         & 45 clicks   & $<$ 120 sec. \\
 LATTE          & 19 clicks    & $<$ 50 sec. \\
\hline
 iDet3D & 2.5 clicks          & $>$ 10 sec. \\
 
\hline
\end{tabular}
\end{adjustbox}
\end{center}
\caption{
    User study comparing the efficiency between manual annotation, LATTE, and our proposed iDet3D.
}
\label{table:user_study}
\vspace{-1\baselineskip}
\end{table}

\section{Concluding Remarks}
In this paper, we propose iDet3D, the first interactive 3D object detector which is capable of detecting numerous multi-class objects within a few clicks.
For effective and efficient 3D detection,
we design NCS to filter out false positive predictions via
negative clicks, and two click propagation modules (DCG and SCP)
to empower user-provided guidance.
Based on our extensive experiments showing the superiority of iDet3D in terms of detection accuracy and efficiency, 
we believe that iDet3D could be a promising option to accelerate data labeling pipelines for LiDAR point clouds.

\paragraph{Future works.}
In this work, iDet3D shows promising results by only analyzing a single frame.
However, most LiDAR scenes are composed of multiple consecutive frames, containing complementary information between each other.
We expect that iDet3D can be further improved in a multi-frame scenario by effectively handling point embeddings
to be aligned between several sequential frames.


{\small
\paragraph{Acknowledgements.}
This work was supported by the National Research Foundation of Korea (NRF) grant funded by the Korea government (MSIT) (No. NRF-2022R1A2B5B02001913) and the Institute of Information \& communications Technology Planning \& Evaluation (IITP) grant funded by the Korea government (MSIT) (No.2019-0-00075, Artificial Intelligence Graduate School Program (KAIST)).}

\bibliography{aaai24}

\end{document}